# Hybrid deep learning methods for phenotype prediction from clinical notes

Sahar Khalafi[1], Nasser Ghadiri[1], Milad Moradi[2]

**Abstract**

**Background and objectives:** Identifying patient cohorts from clinical notes in secondary electronic health records is a fundamental task in clinical information management. The patient cohort identification needs to identify the patient phenotypes. However, with the growing number of clinical notes, it becomes challenging to analyze the data manually. Therefore, automatic extraction of clinical concepts would be an essential task to identify the patient phenotypes correctly. This paper proposes a novel hybrid model for automatically extracting patient phenotypes using natural language processing and deep learning models to determine the patient phenotypes without dictionaries and human intervention.

**Methods:** The proposed hybrid model is based on a neural bidirectional sequence model (BiLSTM or BiGRU) and a Convolutional Neural Network (CNN) for identifying patient's phenotypes in discharge reports. Furthermore, to extract more features related to each phenotype, an extra CNN layer is run parallel to the hybrid proposed model. The extra CNN layer improves the model in identifying different phenotypes. We used pre-trained embeddings such as FastText and Word2vec separately as the input layers to evaluate other embedding's performance in identifying patient phenotypes. We also measured the effect of applying additional data cleaning steps on discharge reports to identify patient phenotypes by deep learning models.

**Results:** We used discharge reports in the Medical Information Mart for Intensive Care III (MIMIC III) database in our experiments to identify patient phenotypes. Experimental results in internal comparison demonstrate that the proposed model achieved significant performance improvement over existing models. In addition, we pre-trained embeddings using FastText and Word2vce on discharge reports in MIMIC III. The enhanced version of our model with an extra CNN layer obtained a relatively higher F1-score than the original hybrid model. We also showed that BiGRU layer with FastText embedding had better performance than BiLSTM layer to identify patient phenotypes.

**Conclusion:** The proposed hybrid model extracts more features than existing methods of extracting patient phenotypes and provides a better F1-score. We show that complementing the proposed hybrid model with an extra CNN in [1] *Department of Electrical and Computer Engineering, Isfahan University of Technology, Isfahan 84156-83111, Iran*
(e-mails: `sahar.khalafi77@ec.iut.ac.ir, nghadiri@iut.ac.ir`)

[2] *Institute for Artificial Intelligence and Decision Support, Center for Medical Statistics, Informatics, and Intelligent Systems, Medical University of Vienna, Vienna, Austria* (e-mail: `milad.moradivastegani@meduniwien.ac.at`)

identifying different phenotypes improves the F1 scores. In addition, eliminating punctuation, numbers, and stop words in discharge reports before training hybrid models resulted in increased model performance.

## 1. Introduction

Electronic Health Records (EHRs) are an integrated set of clinical data related to individual patients. The primary purpose of EHR is to store information for patients' primary care and healing processes [1]. A crucial task in secondary purposes of EHR is correct recognition of patient phenotypes to identify the patient cohort [2]. Identifying patient cohorts in clinical notes is crucial for comparing laboratory tests and treatments within patient cohorts and reducing clinical trials by physicians [3].

An EHR consists of structured (e.g., personal characteristics, diagnostic codes, laboratory results) and unstructured (e.g., discharge reports, nursing notes, radiology reports) data [1]. Most of the information about patient care is unstructured data with no specific format that contains the most valuable clinical details of individual patients [4]. The rapid increase in generating clinical notes in EHR has led to massive amounts of clinical information. The lack of efficient methods and tools for analyzing extensive amounts of information has led to complexity, information loss, uncertainty, and prolonged data analysis. Finally, it leads to difficulty in providing appropriate treatment by physicians [5].

Given these challenges, designing practical tools to extract the intended information from clinical notes has been widely studied in recent years. Clinical notes in EHR have imposed challenges such as lack of a structured framework, spelling errors, poor grammar, complex terms, and ambiguous words. These challenges call for more complex preprocessing to extract helpful information [6, 7].

The Natural Language Processing (NLP) methods based on deep learning models have shown efficiency in analyzing and extracting critical information related to patient phenotypes from clinical notes [8]. An advantage of using deep learning models is learning features without using dictionaries that automatically contain clinical terms. They are also easily generalizable compared with traditional machine learning models [9].

Two commonly used deep learning architectures are CNNs and Recurrent Neural Networks (RNNs), which have also played an essential role in clinical notes classification [10, 11]. CNN could exploit n-grams to encode local semantic features. Variations of RNNs such as GRU[3] [12] and LSTM[4] [13] capture the semantic structure of long-term dependencies.

Patient phenotype identification can be modeled as a binary or multi-label classification problem. In binary classification, every clinical text could be presented with a single phenotype associated with a label. In contrast, in a multi-label classification, each text will be associated with a set of phenotypes. Creating a multi-label classifier that can identify different patient cohorts is more efficient than binary classifiers that only can recognize one patient phenotype. Due to abbreviations, semantic complexities, synonyms, misspellings, and the lack of a clear structure in clinical notes, creating a model for multi-label classification is more complex. The model should be able to learn and identify more variations of features related to different phenotypes.

Deep learning models alone cannot extract all the essential features by investigating the relationships between the input words. For example, CNN cannot record term sequences, and sequential models cannot detect keywords in a given note. Therefore, a model is needed that can take advantage of the combined sequential model and CNN.

In this research, a hybrid model of the bidirectional sequence model with CNN is presented to tackle the clinical notes' challenges and to extract various features to identify patient phenotypes. We also propose an extra CNN with a hybrid model to extract more features related to different phenotypes and improve the hybrid model. Our model uses a bidirectional sequence model to encode bidirectional semantic information. It detects local features from the output with CNN.

For further enhancing the model, an extra CNN identifies local features in parallel, with a combination of adjacent inputs from the embedding layer. The feature maps extracted from both CNNs are fed into the pooling layer to reduce the single value dimensions. Finally, the output of both parallel models is combined and fed to the output layer. The output is a fully connected layer with a sigmoid activation function, containing 1 to 10 neurons. The number of neurons indicates the number of identifiable phenotypes per input.

---

[3] Gated Recurrent Unit
[4] Long Short-Term Memory

The main contributions of this paper are as follows:

- Investigating the effect of data cleansing steps before training the deep learning models on discharge reports
- Proposing a hybrid bidirectional sequence model with CNN in discharge reports to identify different features by considering the semantic structure in long-term dependencies and extracting semantic information from the semantic structure
- Adding an extra CNN in parallel with a hybrid model to identify patient phenotypes
- Evaluating the performance of BiGRU layer versus BiLSTM layer to extract related features to patient phenotypes from discharge reports

Section 2 introduces the clinical notes classification background with deep learning models used in our model. In Section 3, we elaborate on the characteristics of our proposed model in detail. The dataset for experiments, baseline models, and evaluation metrics are covered in Section 4. Section 5 presents experimental parameters and results. Section 6 concludes the paper and points to future research.

## 2. Background

In recent research works, the classification of clinical notes based on deep learning has provided more promising results compared to traditional machine learning models. We use the supervised learning approach to classify clinical notes based on deep learning networks that needs labeling data to train the model.

A key advantage of deep neural networks over traditional predictive models is extracting features from text without human intervention. A widely used deep neural network architecture for classification of clinical notes is *Convolutional Neural Network (CNN)* that is also combined with *sequence models* for some applications. CNN can extract and learn complex features locally [14]. Sequence models, including LSTM and GRU, can understand long-term dependencies and capture semantic structures. Our proposed architecture is based on bidirectional sequence models as Bidirectional GRU (BiGRU) and Bidirectional LSTM (BiLSTM) models.

In a comparative study to extract patient phenotypes from discharge reports in MIMIC III[5], Gehrmann et al. [3] CNN networks performed better at identifying phenotypes than concept extraction methods. No dictionaries and ontologies are required to learn patients' phenotypes and increase the accuracy of the model. They executed CNN on each unit phenotype as binary classification, and used pre-trained embeddings as input layer with word2vec [15] on all discharge reports in the MIMIC III database. Discharge reports in the MIMIC III database were annotated by clinical experts [3], and made available to the public to compare the performance of subsequent work based on ten phenotypes. There is no report on the result of multi-label classification, and the effect of data cleaning steps in CNN. Limiting the space of word search was also a challenge that later addressed by Yang et al. [16]. They expanded the CNN network by adding input at the sentence level that could limit the search space by considering the relationships between sentences. They used local information between words and sentences in extracting patient phenotypes.

Liu et al. [17] presented a CNN model as a binary classifier for predicting readmission of heart failure patients from discharge reports. They use word2vec embeddings trained on PubMed articles as the input layer. The result was better than machine learning models such as a random forest for prediction of patient readmission.

Methods that use CNN to detect features only encode local correlations regardless of the sequence of words and grammar. Therefore, the semantic structure of sequences of words may not be captured by the model. On the other hand, sequence models can capture semantic structure in long-term dependencies. Hashir Khan [18] used a CNN-LSTM model with FastText [19] embeddings as the input layer to predict mortality from discharge reports. The CNN learned semantic relationships with three different filters, and the output entered the pooling layer. Eventually, pooling layers combined and entered LSTM to train temporal relationships. This system was able to achieve better results than disease scoring systems.

Segura-Bedmar and Raez [11] presented a hybrid CNN-GRU model to identify phenotypes from clinical notes. The results were compared with non-hybrid models such as CNN and GRU, where the hybrid model performed better. They also measured the effect of adding a fully connected layer to their network. This layer did not improve the hybrid network's output compared to non-hybrid models.

---

[5] MIMIC III

The advantage of using bidirectional sequence models is that they can encode sentences in both directions, enabling to record semantic and syntactic information [20]. By combining CNN with a bidirectional sequence model, the advantage of both models can be leveraged. To the best of author's knowledge, no existing model for classifying clinical note considers the combination of bidirectional sequence models with CNN. Moreover, the identification of patient phenotypes is modeled as a binary classification problem.

An advantage of our model is the ability to automatically extract and learn patient phenotypes by considering the semantic structure in word order and grammar. It extracts the most salient phrases and clinical concepts as multi-label and binary classification. We also use an extra CNN layer to extract more features related to the patient phenotype. Our model also measures the effect of data cleaning steps on discharge reports in deep learning models output.

## 3. The proposed model

We propose a hybrid of the bidirectional sequence model and CNN to identify patient phenotypes in discharge reports. We also improve the model by an extra CNN to extract more features related to each phenotype.

Pretrained word embeddings such as word2vec or FastText are used as input layers fed into the bidirectional sequence model. The bidirectional sequence model captures the syntactic and semantic structure from discharge reports in both forward and backward directions because it cannot extract keywords in parallel. Therefore, each timestep's hidden state output in the bidirectional sequence model enters the CNN to extract features from the temporal axis. In contrast, for extracting more medical concepts to assign to each patient phenotype, the extra CNN with a top-down filter slider in the input layer produces a feature map from adjacent terms to represent a specific concept.

### 3.1. Preprocessing

Unstructured clinical notes contain imperfections such as spelling errors, abbreviations, lack of grammatical structure, and noise, more complex preprocessing steps are needed to convert texts to usable input before training the model. The discharge reports in the MIMIC III database must also be converted to the appropriate form to identify phenotypes before being used to input different models. We used a two-stage pipeline for preprocessing to improve our model's performance, each stage containing three steps. Details of the preprocessing steps are described in the following.

### 3.1.1. Cleaning the data

To convert text to user input and improve the final model's output, we separately measured the impact of the following three steps of data cleaning on our model. Initially, we converted uppercase letters to lowercase letters. For reducing noise and simplifying the process, we changed all uppercase letters to lowercase. We also removed all punctuation marks, including parentheses, semicolons, periods, etc., as well as numbers. In the second step, we deleted stop words, based on the assumption that stops words do not help in extracting patient phenotypes features. In the third step, we lemmatized the words. All words are converted to their lemma, assuming that lemmatizing words can help reduce the model's complexity and improve performance.

### 3.1.2. Convert discharge reports to the standardized form

The input text must be converted to the numerical form before entering deep neural network models. The following three steps in the pipeline are applied. The first step is tokenizing the input document. Each clinical note is converted into a set of words, and then each word is assigned a unique number. Then, all the notes are padded to the maximum length of notes, since sequence models in deep learning require inputs of the same length to process input documents.

### 3.2. Hybrid bidirectional sequence model with CNN

The architecture of the proposed model is illustrated in Fig. 1. The first layer in our model is the word embedding layer. Word embeddings enter the bidirectional sequence model to save the sequence of information from left-to-right and right-to-left directions. The output from this layer is fed into CNN to learn and locally extract related features to patient phenotype. The feature vectors obtained from the convolution layer enter the global max-pooling layer to reduce the feature vector to a single value and eliminate the noise to get the maximum of the features globally from the feature vectors. Finally, the global maximum of features is entered into the fully connected layer with a sigmoid activation function and one neuron to predict phenotypes individually. In the following subsections, we describe the proposed model in detail.

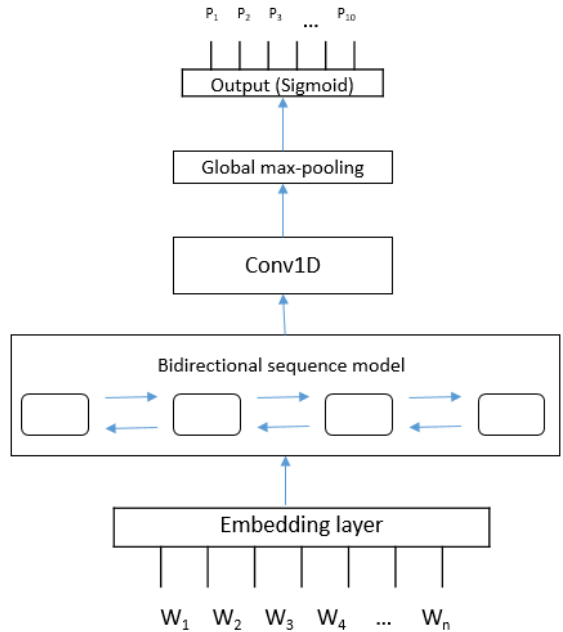

Fig. 1. Schematic of the proposed model architecture to identify different phenotypes from discharge reports. The bidirectional sequence model denotes using the BiLSTM or BiGRU model. The inputs are words $W_{1..n}$ from a clinical note $n$ words. The outputs are the predicted phenotypes $P_{1..10}$.

### 3.2.1. Embedding layer

Word embedding represents the words by vectors containing numbers that make texts comprehensible to machine learning algorithms. A classic representation of words is the one-hot encoding method, which creates vectors with the size of all the words in the input vocabulary. Each element in the vector represents one word in the vocabulary. However, the one-hot encoding method is less used in recent research since the number of dimensions increases by the increased number of words in the vocabulary. The one-hot model also fails to convey any meaning in representations such as semantic relationships between words [21].

Word embedding methods such as word2vec [15] by Google and FastText by Facebook [19] have been recently used to resolve the limitations of traditional word representation methods. These methods rely on neural network algorithms [11]. Word embedding methods can capture semantic similarities between words, as well as morphological, syntactic information, and interrelationships. They can also create meaningful models with smaller and denser dimensions than traditional methods by processing an extensive collection of texts [15].

These algorithms have two basic models, skip-gram and Continuous Bag-Of-Words (CBOW). The CBOW model uses content words as input to predict the target word, while the skip-gram model uses the target word to predict the content words. The architecture of the CBOW and skip-gram models are shown in Fig. 3.

The word2vec embedding recognizes each word as an independent unit and creates a vector for every word. FastText deals with each word as n-grams of characters and creates a vector for every n-gram such that the sum of the vectors represents one word. This enables the FastText model to perform better when facing new words and understand the word structure [22].

For evaluating the performance of different embeddings, we pre-trained the embeddings with FastText and word2vec on discharge reports in the MIMIC III database with the skip-gram model as an input layer to the proposed model. Each word in the discharge reports is mapped to its corresponding numeric vector in the word embedding layer. The output of this layer is a two-dimensional matrix. We separately applied the data cleaning steps on discharge reports before training the embeddings.

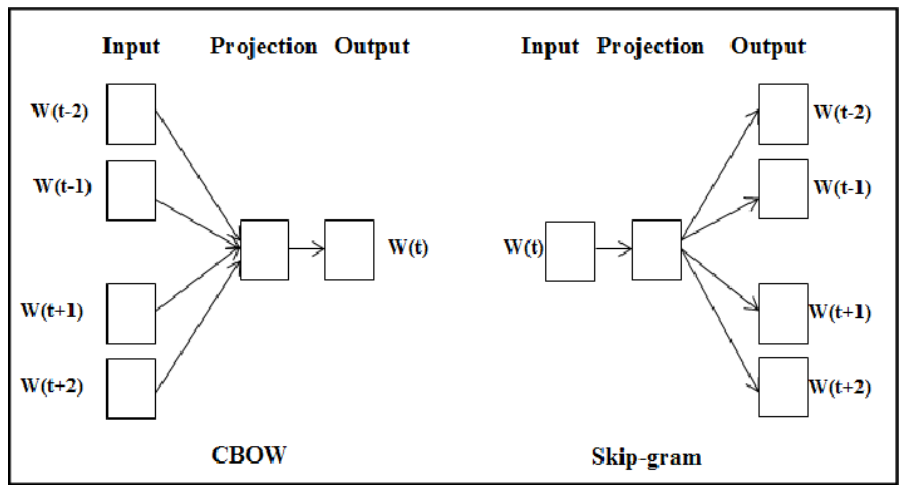

Fig. 3. The architecture of the CBOW and skip-gram models [23]. CBOW estimates the probability of the target word given the context words, while skip-gram estimates the probability of the context words given the target word.

### 3.2.2. Bidirectional sequence models

A primary application of the RNN architecture is in learning from sequential data, where the current state depends on the previous state's input token and activation function. This feature captures the sentence structure and long-term dependencies in the discharge reports.

The RNN uses gradients to update the network's parameters. In backpropagation, the gradients are propagated from the end of the network to the beginning. As the network grows, the gradients become exponentially smaller, causing the vanishing gradients problem. Therefore, to solve the vanishing gradients problem, sequence models such as LSTM and a simplified version of LSTM called GRU have emerged. They solve the vanishing gradients problem by adding long-term memory and can store long-term dependencies.

A central element in LSTM is the *memory cell*, which uses three gates in the LSTM architecture, including forget and input gates, to update the memory content [13]. The *forget gate* controls old information to determine how much memory information was erased in the previous state. New input information is controlled by the *input gate*, which contains two layers as follows:

1- The first layer shows which part of the memory cell's input information is updated by applying a sigmoid activation function.

2. The second layer decides which input elements are preserved to the memory cell (by generating a vector of previous memory cell values with the tangent activation function).

The output of each gate generated by applying the sigmoid activation function is a number between zero and one. A value of zero means clearing information and not sending it to the next state. In contrast, a value of one refers to sending all information in the current state to the next state.

The GRU network is a simplified version of the LSTM architecture; by merging LSTM gates, GRU gates have been reduced to update and reset gates [12]. The update gate determines how much of the current information, previous information, or a combination of the two will be used in state *t* without overwriting and deleting the information. This feature maintains the long-term dependency of the information in the previous states and affects later states using memory. The reset gate specifies how much information is deleted from previous steps in the current step.

The sequence models only retain input information they have already seen, so the input information preserves past and future data. We use bidirectional sequence models, as shown in Fig. 4, consisting of two forward and backward sequence models, which can obtain sequential text information from both directions. It helps the model to understand the semantic information contained in the text better and ultimately improves the model's output.

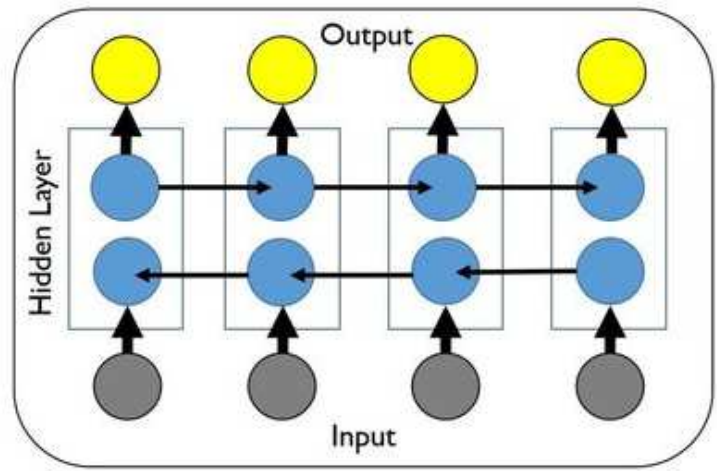

Fig 4. Bidirectional sequence model architecture [24].

The second layer in our proposed model is the *bidirectional sequence model*. The embedding layer's output, including word vectors, is entered into the bidirectional sequence model separately to capture semantic dependencies between words in two directions. At time *t,* a word vector enters the bidirectional sequence model. This word depends on the previous word at time *t-1*. The output generation process at each time step is shown in Fig. 5. At each timestep, the output $h_t$ is calculated using the current input $x_t$ and all previous outputs $h_{t-1}$ by the tangent activation function. The output of each timestep is entered into CNN to extract semantic features from the long-term dependency between words.

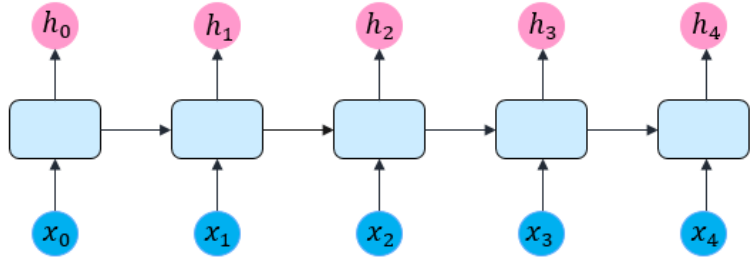

Fig 5. Word vectors in a sequence model. The sequence model refers to LSTM or GRU models. Inputs at each time step are denoted by $x_i$, and $h_i$ means outputs at each time step.

### 3.2.3. Conv1d

The CNN architecture is a commonly used deep neural network for feature extraction, feature selection, and pattern classification. CNN has provided significant results in computer vision and image processing since its advent [25].

In recent years, the potential of CNN in the NLP domain for extracting local features and recognizing salient phrases from the text has gained attention. In addition, they have achieved excellent performance in text classification. CNN's are composed of different layers, including the convolutional, pooling, and fully connected layers [8].

The convolutional layers consist of a set of filters, also called *kernels*. The output of each timestep of the bidirectional sequence model is fed to the *conv1d* layer. After that, a convolution operation is performed on the temporal axis. Fig. 6 shows how the features are extracted through applying the convolution operation by the conv1d layer of the temporal axis. Fig. 7 shows how the convolution operation works. The filters move like a sliding window from top to bottom across the temporal axis and perform the convolutional operation, which is the element-wise product between the local area and the filter. Finally, all the production values are added together, and the filtered output is obtained.

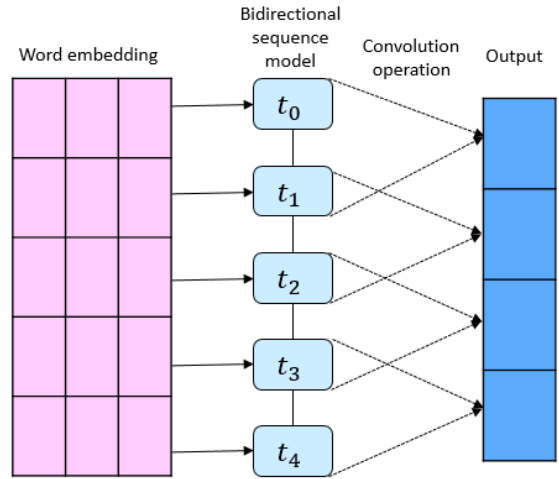

Fig.6. Schematic of feature extraction through applying the convolution operation by the conv1d layer of the temporal axis.

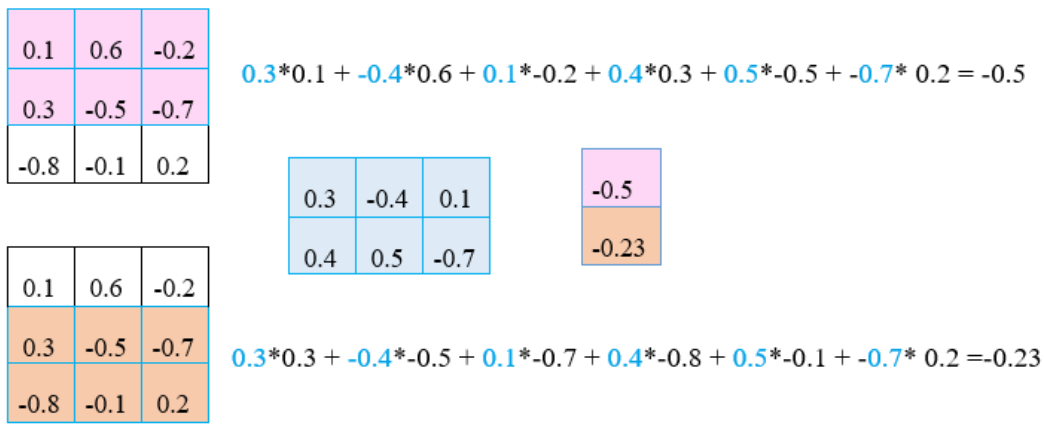

Fig. 7. Convolution operation to extract local features. The blue rectangle (middle rectangle) is a filter that moves from top to bottom on the temporal axis (left rectangles) with a stride of one and performs a convolution operation.

Filters initially have random weights that are updated during backpropagation to achieve acceptable performance. Each feature extracted from each selected filter enters the pooling layer. The pooling layer uses a static activation function while preserving the most salient phrases to reduce the feature space's size resulting from convolutional operations. The pooling layer reduces the number of parameters and calculations. The pooling layer has two models: max-pooling and average-pooling, which refer to selecting the top feature and averaging features from feature vectors. Since not all the extracted features are equally important, using the max-pooling model works better than the average-pooling. In this study, we used the one-dimensional global max-pooling layer to select the global features; this layer's output is used as the input of the next layer [26].

From the CNN point of view, feature vectors can be a semantic representation of the input and enter into the output layer, a fully connected layer. Output layer uses the sigmoid activation function on its input features; its output consists of a vector containing predicted labels as $\hat{y} = [\hat{y}_1, \hat{y}_2, \hat{y}_3, \dots, \hat{y}_i]$ where $Y^i$ refers to $i_{th}$ phenotype labels.

### 3.3. The enhanced version of the hybrid model

The previous section presented our hybrid proposed model to identify patient phenotypes from the clinical note. Further improvement could be achieved to extract more medical concepts to assign each phenotype. It enables identifying phenotypes as a result of multi-label classification. For this purpose, we added a CNN, which works in parallel with the hybrid model. It provides more information to the model for extracting phenotypes as a multi-label classifier.

The architecture of the enhanced proposed model is illustrated in Fig. 8. The extra CNN works similar to the CNN in our hybrid model as described in Section 3-2-3. The difference is that the convolution operation is performed on the output of the embedding layer. The filters' width in convolution operation is equal to the word embedding size, and its height is adjustable. We used a one-dimensional convolution for extracting semantic features from each filter. The filters play the role of n-grams in extracting salient phrases related to patient phenotypes. As shown in Fig. 9, different filter sizes are equal to different n-grams [26].

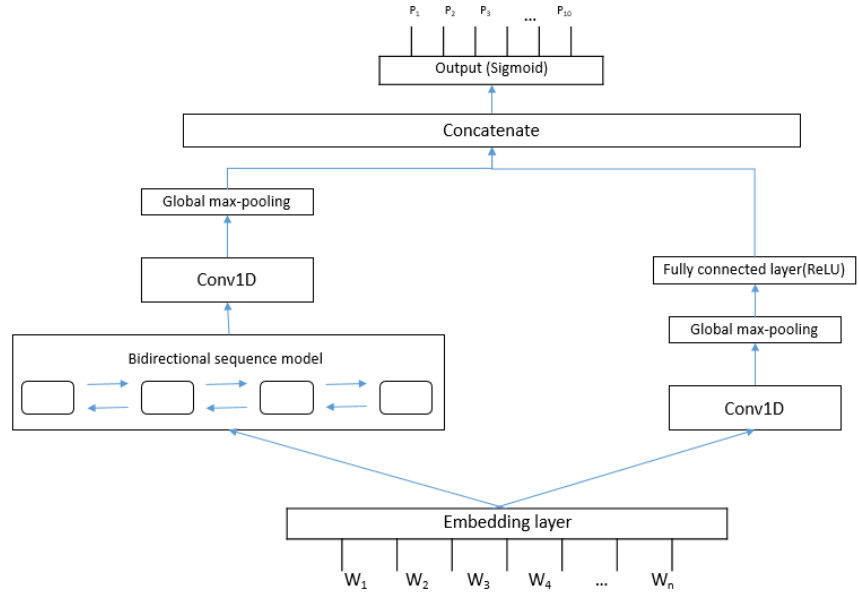

Fig. 8. The architecture of the enhanced version of the proposed model. The bidirectional sequence model denotes using the BiLSTM or BiGRU models. The input includes a clinical note with $w_n$ words, where $n$ is the number of words in a clinical note. The output includes the predicted phenotypes.

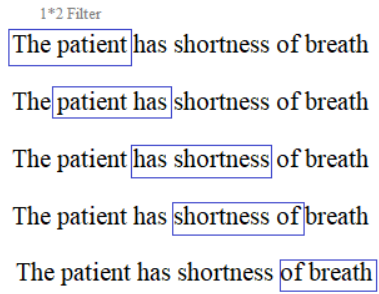

Fig. 9. One-dimensional convolution operations on a clinical note. The filter plays the role of 2-gram in this example.

The feature vectors obtained from the hybrid model and extra CNN will enter the global max-pooling layer to reduce the dimensions and eliminate the noise to get the maximum of the features globally from the feature vectors. The output of the global max-pooling layer of the non-hybrid CNN is fed into a fully connected layer with a ReLU activation function. The output of the hybrid model is concatenated with the output of non-hybrid CNN. Finally, the features enter a fully connected layer with a sigmoid activation function for predicting patient phenotypes.

### 3.4. Performance evaluation of bidirectional sequence models

In this paper, we intend to compare the performance of bidirectional sequence models such as BiLSTM and BiGRU in identifying patient phenotypes. Since GRU is a simplified version of LSTM, it is expected that the execution time of the model with this layer will be higher and have a performance close to LSTM.

## 4. Experimental evaluation

### 4.1. Dataset

The MIMIC III database is an extensive database containing EHRs of adult patients in the intensive care unit of Beth Israel Deaconess Medical Center between the years 2001 and 2012 [27].

The NOTEEVENTS table is a data table in the MIMIC III database that contains open-source clinical notes, including nursing, medical, radiological, and discharge reports for individual patients. We used annotated discharge reports provided by Gehrmann et al. [3] from the NOTEEVENTS table to evaluate our proposed method. Discharge reports contain complete information about patient phenotypes [3]. Each discharge report consists of sections such as date of admission, date of discharge, date of birth, gender, allergy, primary complaint, major surgery or invasive procedure, medical history, physical examination, etc.

We used Gehrmann et al. [3] annotation to train and test our models. Fig. 10 shows the distribution of 10 annotated phenotypes in the discharge reports. Since the number of positive samples is less than the number of negative samples, the dataset is imbalanced. The total number of annotated discharge reports is 1610 samples. A binary value for 10 different phenotypes is assigned to every report. The value of one for each phenotype refers to the presence of that phenotype in the relevant discharge report. The dataset was segmented so that 20 percent of the total discharge reports were considered test data, another 20 percent for validation, and 60 percent for model training.

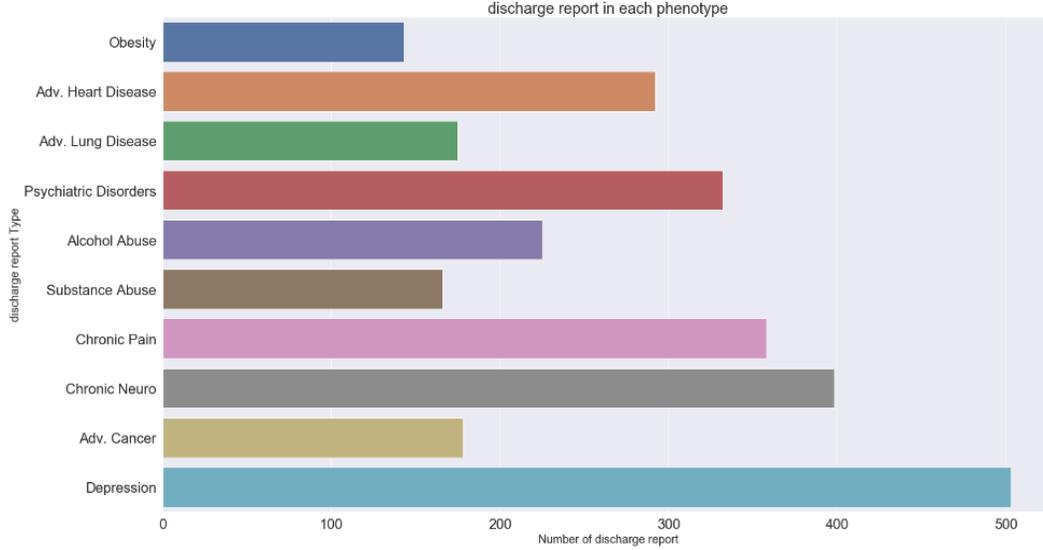

Fig. 10. Distribution of phenotypes in discharge reports

## 4.2. Evaluation metrics

Since our data are unbalanced and most of the data are negatively labeled, the accuracy metric is not sufficient for the analysis of the model. The accuracy metric divides the number of correctly predicted cases by the total number of samples. This is not reliable for unbalanced data. Therefore, we used F1-score to evaluate the output of the binary classification models. F1-score is equal to the harmonic mean of recall presented by Eq. (2) and precision presented by Eq. (3) metrics. The F1-score formula is defined by Eq. (4).

$$Recall = \frac{TP}{TP + FN} \qquad (2)$$

$$Precision = \frac{TP}{TP + FP} \qquad (3)$$

$$F1\text{-}score = \frac{2(Recall * Precision)}{Recall + Precision} \qquad (4)$$

where $TP$, and $FP$ are positive samples that have been correctly and incorrectly classified, respectively. $TN$ and $FN$ denote negative samples that have been correctly and incorrectly classified, respectively. We used F1-micro and F1-macro scores for the final evaluation of the multi-label classification models.

$$Micro\text{-}avg\ Recall = \frac{\sum_{c=1}^{C} TP_c}{\sum_{c=1}^{C} TP_c + FN_c} \qquad (5)$$

$$Macro\text{-}avg\ Recall = \frac{\sum_{c=1}^{C} Recall_c}{C} \qquad (6)$$

$$Micro\text{-}avg\ precision = \frac{\sum_{c=1}^{C} TP_c}{\sum_{c=1}^{C} TP_c + FP_c} \qquad (7)$$

$$Macro\text{-}avg\ precision = \frac{\sum_{c=1}^{C} Precision_c}{C} \qquad (8)$$

$$Micro\text{-}avg\ F1\text{-}score = \frac{2(Micro\text{-}avg\ Recall * Micro\text{-}avg\ Precision)}{Micro\text{-}avg Recall + Micro\text{-}avg\ Precision} \qquad (9)$$

$$Macro\text{-}avg\ F1\text{-}score = \frac{2(Macro\text{-}avg\ Recall * Macro\text{-}avg\ Precision)}{Macro\text{-}avg\ Recall + Macro\text{-}avg\ Precision} \qquad (10)$$

C denotes the number of labels (phenotypes) available for classification; Macro-average calculates the metrics independently for each label and then takes the average. Micro-average calculates the aggregation of contributions of all labels to compute the averaged metric.

### 4.3. Baseline models

This section describes the baseline models used to compare the output of multi-label classification.

- CNN: A convolution neural network proposed by Gehrmann et al. [3] to identify patient phenotypes from discharge reports.
- ws-CNN: A convolution neural network with three different filter sizes with a combination of word and sentence level embeddings proposed by Yang et al. [16] to identify patient phenotypes from discharge reports.
- Sequence models including LSTM and GRU that record long-terms dependencies unilaterally, followed by a pooling layer.
- Bidirectional sequence models such as BiLSTM and BiGRU capture semantic structure in both directions, followed by a pooling layer.
- CNN+LSTM: The hybrid model proposed by Khan [18] so that the output of the CNN layer enters the LSTM layer to predict mortality from discharge reports.

- Bidirectional sequence model + CNN (S-Conv-nm) is the proposed model that records long-term dependencies in two directions. The bidirectional sequence model is fed to the CNN to extract local features with a pooling layer.
- Bidirectional sequence model + CNN – CNN (IS-Conv-nm) is the improved version of the proposed model. Bidirectional sequence model followed by a CNN and pooling layer and in parallel, an extra CNN followed by a pooling layer, dropout, and a fully connected layer. Finally, the outputs of the two CNN are merged.

## 5. Results and discussion

In the following, we first fine-tune hyperparameters of the proposed model. Finally, we evaluate the proposed model with the baseline models.

### 5.1. Hyperparameters setting

We optimize the hyperparameters' value using the 5-fold-cross-validation and report the best values in Table 1. All models were trained using an early stopping based on no change in validation losses after three patience. The layers used in all models are one-dimensional.

Fig. 11 shows that the best filter size for feature extraction on CNN is 2, which plays the role of 2-grams an embedding layer. Fig. 12 shows the effect of the number of hidden units of CNN and BiLSTM layers in the proposed model on the F1 score. As the number of units increases from 64 to 512, the F1-score improves. It is noteworthy that increasing the number of neurons from 512 does not significantly increase the F1 score but increases the computational cost and, ultimately, the model's training time. This study considers the number of hidden units for CNN, LSTM, and GRU to be 512.

All the experiments presented in this article were performed on a Google Colab platform with GPU-based computations. The models were implemented using Python, TensorFlow, and Keras.

Table. 1. Hyperparameter setting in proposed models. The results are obtained from 5-fold cross-validation.

| Parameter | Value |
|---|---|
| Word vector dimension | 300 |
| Word vector model | Skip-Gram |
| Loss function | Binary Cross entropy |
| Optimizer | Adam |
| Batch size | 64 |
| Pooling Method | Global max-pooling |

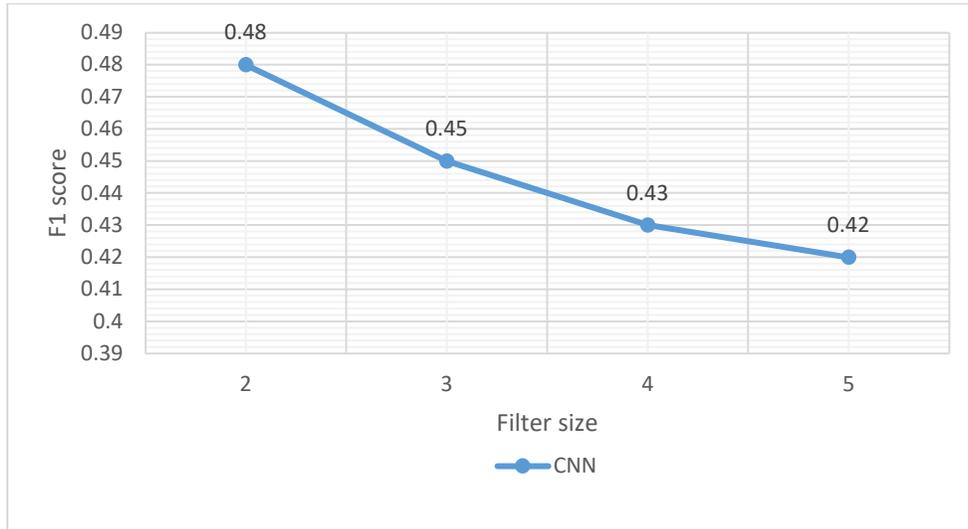

Fig. 11. The impact of different values of filter size on the CNN. Variations of macro F1-scores with different filter sizes in the output of CNN as a multi-label classification to identify different phenotypes. The results are obtained from 5-fold cross-validation.

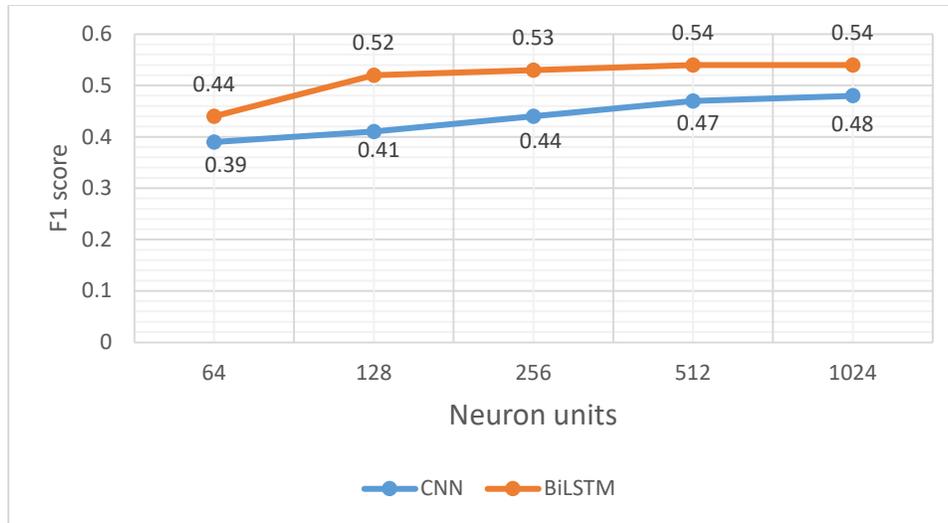

Fig. 12. The impact of the number of hidden units on the macro F1-score of CNN and BiLSTM layer in the proposed model as multi-label classification in identifying different phenotypes. The results are obtained from 5-fold cross-validation.

## 5.2. Data cleaning experimental results

We report multi-label classification results to identify phenotypes by applying each data cleaning step before training word2vec embeddings on discharge reports in Fig. 13. The results are reported for running all models ten times and averaging F1 scores. The first step in cleaning data is to remove punctuation and numbers. The second step is to remove stop words. The third step is to lemmatize words, every step following the previous one.

As shown in Fig. 13, applying the first and second data cleaning steps improves the F1 score of the models except for BiLSTM. The BiLSTM model considers the sequence of dependencies to record semantic information. Therefore, the more complete the input information, the more accurate the output. Lemmatizing words enhance the CNN's output by 0.03 percent because CNN does not consider long-term dependencies and extracts local features.

In the following, all evaluations of the proposed model have been performed by applying the first and second data cleaning steps on discharge reports.

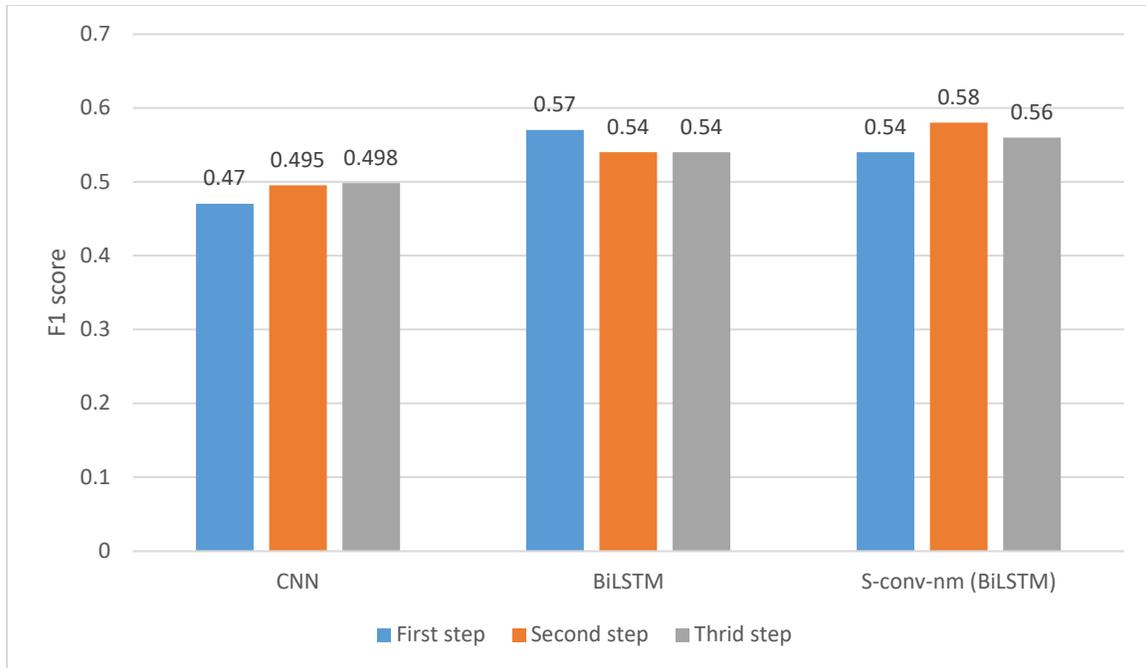

Fig. 13. Effect of applying data cleaning steps in discharge reports before training models. Data cleaning includes three steps, the first step eliminating punctuations with numbers, the second step eliminating punctuations with numbers and stop words, the third step eliminating punctuations with numbers, stop words, and lemmatizing words.

### 5.3. Comparison of Basic deep learning models with the proposed hybrid model

Comparative results are shown in Table 2. Following the comparison of our results, it was shown that the proposed model has a significantly better performance than the experiment models, taking advantage of the bidirectional sequence model and CNN.

Our experiments showed that using ws-CNN with the combination of sentences and word-level input does not necessarily improve model performance. CNN performed better than the ws-CNN and LSTM models in identifying patient phenotypes from discharge reports.

Because the discharge reports have grammatical complexities, LSTM models that record information sequences in one direction reduce the model's performance. CNN extracts the essential n-gram information and performs better than LSTM. The BiLSTM model can capture information in two directions to improve F1-scores compared to the LSTM model.

In hybrid models, the results of CNN+LSTM show that complicating deep learning models does not necessarily improve the final performance of the model. The CNN+LSTM hybrid model first extracts semantic information using CNN and then extracts the dependencies between features with LSTM. The CNN+LSTM hybrid model is less efficient in identifying patient phenotypes than the other models evaluated.

In the proposed model, we added a CNN to the bidirectional sequence model. In the proposed model advantages of both models can be exploited by capturing long-term sequences while extracting local features from these sequences. The proposed model results in increasing the F1 scores. As shown in Table 2, we found that using the BiGRU layer instead of BiLSTM in the hybrid model could significantly improve the final model's performance in identifying patient phenotypes as multi-label classification.

Table 2. Comparison F1-score results to identify patient phenotypes as multi-label classification with the proposed model and the comparison models. The results are presented for different word embeddings. The average time shows the average execution time of the models in each epoch. The proposed model with a bidirectional sequence model with CNN is specified by an asterisk (*) sign. The highest score in each column is shown in bold type.

| Model | word2vec embedding | | FastText embedding | | Avg.time |
|---|---|---|---|---|---|
| | F1 micro | F1 macro | F1 micro | F1 macro | |
| CNN (2018) | 0.590 | 0.546 | 0.604 | 0.588 | 9s |
| ws-CNN (2020) | 0.557 | 0.527 | 0.569 | 0.536 | 11s |
| LSTM | 0.577 | 0.525 | 0.575 | 0.505 | 57s |
| BiLSTM | 0.610 | 0.570 | 0.581 | 0.532 | 93s |
| BiGRU | 0.593 | 0.563 | 0.620 | 0.588 | 120s |
| CNN+LSTM (2019) | 0.428 | 0.391 | 0.401 | 0.337 | 60s |
| S-conv-nm *(BiLSTM) | 0.610 | 0.581 | 0.582 | 0.549 | 96s |
| S-conv-nm *(BiGRU) | **0.620** | **0.589** | **0.626** | **0.608** | 121s |

## 5.4. Comparison of proposed hybrid model with improved version of proposed hybrid model

Comparing the improved version of the proposed model with the hybrid model, as shown in Table 3, we found that since multi-label classification is more complex than binary classification, it is necessary to extract more features at different levels to identify each of patient phenotype. More features can be extracted by adding an extra CNN that runs in parallel with the hybrid bidirectional sequence model with CNN. The features extracted from each layer provided more information to the final model, which improves the final performance.

Table 3. Comparison of F1-score results to identify different phenotypes with the proposed model and improve the proposed model version. The results are presented for different word embeddings. The average time shows the average execution time of the models in each epoch. The hybrid model with the bidirectional sequence model with CNN is the proposed model, and hybrid models with two CNN improve versions of the proposed model. The highest score in each column is shown in bold type.

| Model | Word2ve embedding | | FastText embedding | | Avg.time |
|---|---|---|---|---|---|
| | F1 micro | F1 macro | F1 micro | F1 macro | |
| S-conv-nm (BiLSTM) | 0.610 | 0.581 | 0.582 | 0.549 | 96s |
| IS-conv-nm (BiLSTM) | 0.621 | 0.595 | 0.617 | 0.594 | 97s |
| S-conv-nm (BiGRU) | 0.620 | 0.589 | 0.626 | 0.608 | 121s |
| IS-conv-nm (BiGRU) | **0.636** | **0.612** | **0.630** | **0.614** | 122s |

We also measure the performance of the proposed model in identifying patient phenotypes as binary classification with word2vec embeddings and report the results in Table 4. The proposed model performed better than compared models in identifying phenotypes as binary classification.

Table 4. Comparison of F1-scores results to identify patient phenotypes as binary classification with CNN and proposed models with word2vec embeddings on evaluation corpus. The proposed model with bidirectional sequence model with CNN and improved version of the proposed model with extra CNN are specified by an asterisk (*) sign. The highest score in each row is shown in bold type.

| Model | CNN 2018 | CNN+LSTM 2019 | ws-CNN 2020 | *S-conv-nm | *IS-conv-nm |
|---|---|---|---|---|---|
| Adv. Heart Disease | 0.75 | 0.74 | 0.74 | **0.78** | 0.76 |
| Adv. Lung Disease | 0.62 | 0.65 | 0.59 | **0.76** | 0.73 |
| Adv. Cancer | 0.74 | 0.76 | 0.75 | 0.82 | **0.84** |
| Chronic Pain | 0.57 | 0.61 | 0.69 | 0.66 | **0.70** |
| Substance Abuse | **0.83** | 0.68 | 0.69 | 0.79 | 0.79 |
| Obesity | **0.97** | 0.62 | 0.84 | 0.79 | 0.79 |
| Psychiatric disorders | 0.83 | 0.57 | 0.83 | **0.85** | **0.85** |
| Depression | 0.84 | 0.61 | 0.85 | **0.88** | **0.88** |
| Chronic Neurologic | 0.69 | 0.52 | 0.62 | 0.66 | **0.71** |
| Alcohol Abuse | **0.81** | 0.60 | 0.80 | 0.77 | 0.78 |

## 5.5. Evaluating the performance of bidirectional sequence models

As shown in Fig.14, FastText embeddings in models containing the BiGRU layer performed more efficient than word2vec embeddings. In contrast, in models containing the BiLSTM layer, word2vec embeddings performed better than FastText. word2vec and FastText embeddings had almost similar results due to multi-label classification in the improved hybrid model. Finally, it can be obersved that hybrid models with the BiGRU layer provides higher performance than the hybrid models with the BiLSTM layer in identifying phenotypes as multi-label classification from discharge reports.

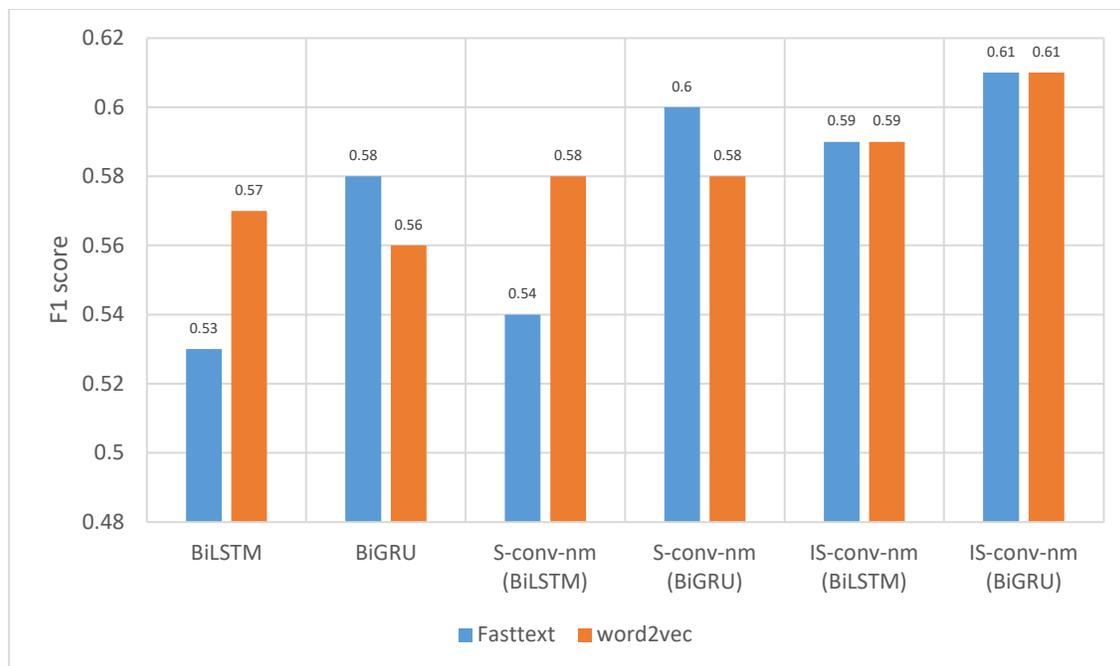

Fig. 14. Comparison F1 score of models as a multi-label classification to identify different phenotypes with FastText and word2vec embeddings

## 6. Conclusion

Using NLP and deep learning in this paper, we identified patient phenotypes in single and multiple forms. The experimental results showed that the proposed combined and improved model could extract more features to identify phenotypes better than the compared models. Experiments with data cleaning steps on discharge reports have shown that removing punctuation, numbers, stop words, and lemmatizing words improve CNN output. On the other hand, eliminate stop words and lemmatizing led to the performance reduction in the bidirectional sequence model. In hybrid models, only removing punctuation, numbers, and stop words improves the model's final performance. We also showed that word2vec and fastText embeddings in the proposed model(improved version) have almost the same performance. In the case of models containing the BiLSTM layer, word2vec embeddings performed better than fastText embeddings. In the models including the BiGRU layer, fastText embeddings performed better.

In future works, we will use context-aware embeddings such as Bidirectional Encoder Representations from Transformers (BERT) [28] as the input layer. BERT can improve deep learning models' performance. Context-aware embeddings generate a vector for each word by considering how the word appears; word vectors dynamically change with different contexts. Moreover, we could use phenotype-related concepts in clinical terms dictionary input such as UMLS to combine with unstructured data. Combined phenotype-related concepts with unstructured data will help to increase the performance of extracting phenotypes.